\documentclass[10pt]{article}
\usepackage[letterpaper]{geometry}
\usepackage{hicss}
\usepackage{times}
\usepackage[none]{hyphenat}
\usepackage{url}
\usepackage{latexsym}
\usepackage{indentfirst}
\usepackage{graphicx}
\graphicspath{{images/}}
\usepackage[
    natbib=true,
  ]{biblatex}
\newif\ifshowauthorcomments

\showauthorcommentstrue

\usepackage{enumitem}
\usepackage{relsize}

\title{Mates2Motion: Learning How Mechanical CAD Assemblies Work\\[0.2em]\smaller{}\smaller{}Presented at the ICML 2022 Workshop on Machine Learning in Computational Design}
\author{James Noeckel \\ University of Washington \\ {\underline{jamesn8@cs.washington.edu}} \And Benjamin T. Jones \\ University of Washington \\ {\underline{benjones@cs.washington.edu}} \And Karl Willis \\ Autodesk Research \\ {\underline{karl.willis@autodesk.com}} \AND Brian Curless\\ University of Washington \\ {\underline{curless@cs.washington.edu}}
\And Adriana Schulz \\ University of Washington \\ {\underline{adriana@cs.washington.edu}} }

\date{}

\begin{document}
\maketitle
\begin{abstract}
We describe our work on inferring the degrees of freedom between mated parts in mechanical assemblies using deep learning on CAD representations. We train our model using a large dataset of real-world mechanical assemblies consisting of CAD parts and mates joining them together. We present methods for re-defining these mates to make them better reflect the motion of the assembly, as well as narrowing down the possible axes of motion. We also conduct a user study to create a motion-annotated test set with more reliable labels.
\end{abstract}

\section{Introduction}
\label{sec:intro}

Understanding how mechanical assemblies move is important for applications ranging from simulating mechanical behaviour to robotic manipulation. While abundant models of mechanical assemblies exist in the wild, they largely consist of static geometry devoid of motion information. However, since mechanical assemblies typically start their lives in  Computer-Aided Design (CAD) software, most of these models \textit{were} once annotated with critical information for understanding how articulated assemblies function. This is because, in addition to the geometry, CAD systems allow designers to specify degrees of freedom between assembled parts. The goal of this project is to recover this information.

 Typically, CAD systems represent motion in an assembly using {\it mates}, which specify the degrees of freedom between pairs of parts. However, in most large repositories of 3D models, this motion information is absent. The primary reason is that each CAD system has its own internal and proprietary method for keeping track of mate information. CAD models are generally exported and exchanged in a purely geometric representation, called a B-Rep (boundary representation). B-Reps are the common format in large repositories of man-made shapes~\cite{grabcad,koch2019abc,willis2021fusion,jones2021sb}, and while B-Reps describe the geometry of parts in an assembly, they do not include information about their mechanical degrees of freedom. This work therefore proposes to use B-Reps as the input format to our inference problem.

While prior work has addressed the problem of inferring motion from static assemblies~\cite{hu2017learning, wang2019shape2motion, yan2020rpm}, they do not work directly with CAD assemblies (B-Reps), using instead geometric datasets (point clouds or meshes) that have been hand-annotated. The drawback of these approaches is that they are restricted to specific classes of common objects which are known to exhibit motion, rather than working on arbitrary mechanisms in the wild. Recently, large repositories of CAD assemblies have been made public that include detailed information about how parts are mated together and move~\cite{jones2021sb,willis2021joinable}. These collections include a large variety of mechanical parts using rich CAD representations that have the potential to enable inference beyond specific object classes. This work aims to address the fundamental question: Can we learn to infer motion in general from collections of CAD assemblies?  

\begin{figure}
    \centering
    \includegraphics[width=7cm]{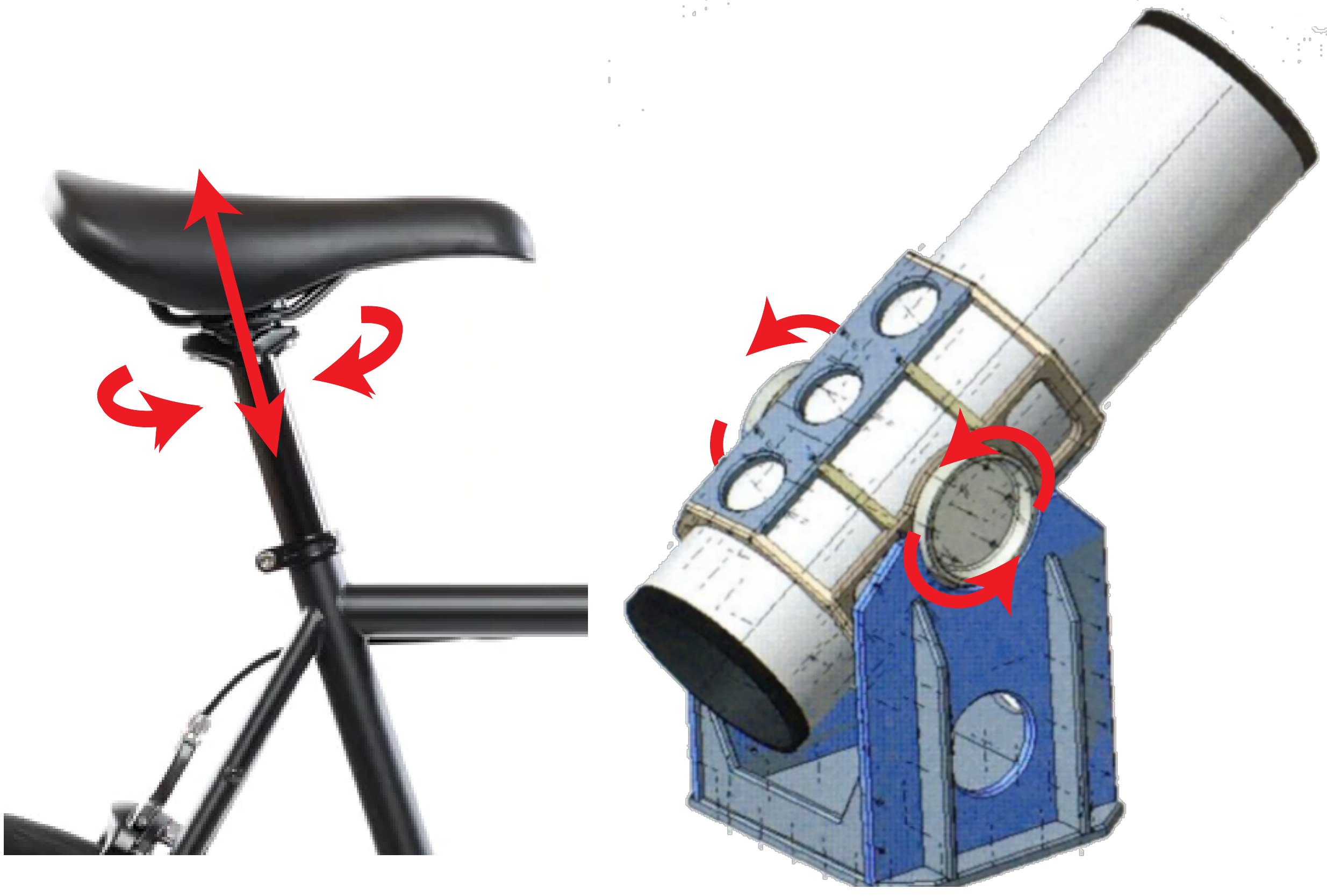}
    \caption{Sources of ambiguity. Left: A bicycle seat could be classified as sliding/rotating/fixed depending on which aspects of the bicycle's operation the {\it user intends} to model. Right: A single mate on either side of the telescope is sufficient to represent the motion creating {\it motion to mate} ambiguity.}
    \label{fig:dataambiguity}
\end{figure}

While CAD assembly repositories are a valuable source of real-world assemblies, they also present several challenges to learning. The user's intent when creating an assembly affects which type of motion to use, or whether to annotate motion at all, leading to ambiguous or missing motion labels. Even a deterministic set of motions for an assembly has several equivalent ways to represent it using mates, 
leading to conflicting signal for learning-based methods seeking to understand just the motion through the mates (Figure \ref{fig:dataambiguity}.)

In this work we take the first step at addressing these challenges to learn how mechanical assemblies work from collections of CAD mates. Our key contributions are:
\begin{enumerate}[noitemsep]
    \item A Dataset of moving assemblies for learning, with filters and a modified representation to mitigate the errors and ambiguity found in raw assemblies
    \item A user-annotated validation set
    \item Baselines for the mate prediction task.
\end{enumerate}


\section{Background}
\label{sec:related}

\paragraph{Learning-Based Motion Inference from Static Geometry}
Recent work has taken a data-driven approach to motion inference. \citet{hu2017learning} use metric learning to query similar articulated pairs from a database. \citet{wang2019shape2motion} train separate motion proposal and optimization networks on point clouds to segment parts and infer motion as a motion \textit{type} and \textit{axis} for each part. This motion representation captures the rigid motions exhibited in mechanical objects, and is what we use in this work. \citet{yan2020rpm} instead infer per-point displacements, allowing more general part motions to be represented. \citet{xu2022unsupervised} avoid needing joint annotations by learning similarity transformations within a semantic object category. All of these works rely on semantic object categories to make learning tractable. We seek to leverage the richer information of B-Rep geometry to remove reliance on semantic knowledge.

\paragraph{Learning on CAD Data}

Large repositories of CAD formatted geometry and assemblies have long been available~\cite{grabcad}, and there has been an explosion of large curated~\cite{koch2019abc} and annotated~\cite{seff2020sketchgraphs,willis2021fusion,jones2021sb} datasets. B-Reps have a natural graph representation, which has inspired several graph neural networks for B-Rep modeling and classification tasks~\cite{cao2020graph,jayaraman2021uvnet,meltzer_uvstyle-net_2021,lambourne2021brepnet}. Two works address B-Rep assemblies; \citet{jones2021sb} predict part alignment and joint type given rough locations on pairs of parts and provides the dateset used in this work, while \citet{willis2021joinable} predict alignment axes and offsets for part pairs, but not joint type. In contrast, this work operates at the assembly level, predicting part connectivity, connection axes, and joint type. Our inputs are already positioned relative to one another, so offset prediction is not required.

\section{Dataset}\label{sec:data}

\begin{figure}
    \centering
    \includegraphics[width=5cm]{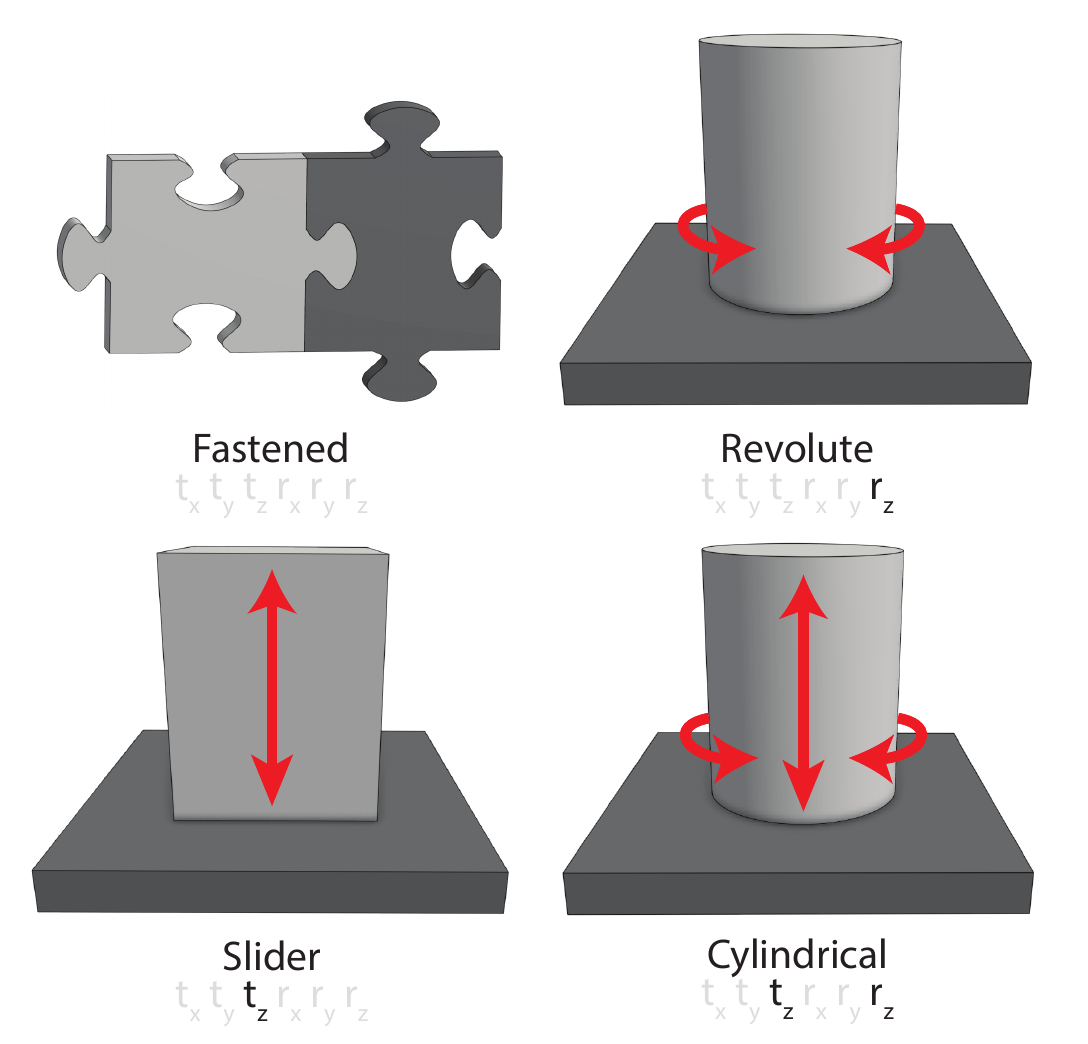}
    \caption{The four mate types and their DoFs.}
    \label{fig:matetypes}
\end{figure}

We build upon the dataset of \citet{jones2021sb}, containing 125,133 CAD assemblies created on the Onshape platform. 
There are several key issues with learning directly from this dataset:
\begin{enumerate}[noitemsep]
    \item \textit{Noise}: Many assemblies are incomplete, or learning/practice material.\label{item:0}
    \item \textit{Motion-to-mate ambiguity}: the same motion can be represented by many different choices of mates.\label{item:1}
    \item \textit{User intent ambiguity}: The correct mate/motion may depend on what the user considered important.\label{item:2}
    
\end{enumerate}

Making use of the vast amount of CAD assemblies at our disposal necessitates correcting some of the ambiguity inherent in the dataset. We address this problem in three ways. First, we propose a set of automatic methods for filtering the assemblies to remove noise. While these heuristic-driven methods cannot ensure that all non-plausible mates are resolved, these methods are able to remove a large number of models that are not physically plausible. Second, we propose a novel method to automatically remove motion-to-mate ambiguity by redefining mates, addressing problem (2). Finally, we create a validation set that is free of noise (1) and user intent ambiguity (2) by running a user study over a hand-selected portion of the dataset.

\paragraph{Heuristic Filters} To mitigate problem (\ref{item:0}), we apply several heuristic filters based on our understanding of the domain. We first filter for assemblies that have at least one moving part to fit our task. To account for designs that are incomplete, we filter out all assemblies with disconnected pieces, i.e. separately connected sub-assemblies rather than a single connected assembly. We further remove what we call \textit{compound mates}, in which multiple, possibly conflicting, mates are defined between the same pair of parts, rendering the motion invalid. Finally, we remove spherical and pin-slot mates, which are seldom used, and parallel and planar mates, which are seldom used correctly. Our assemblies consist of four common mechanical joints: Fastens, Revolutes, Sliders, and Cylindricals (see Figure \ref{fig:matetypes}). These filters leave us with 13,957 assemblies, which we call the \textit{cleaned} set.

\paragraph{Removing Motion-to-Mate Ambiguity} 
To address (\ref{item:1}), we augment our dataset so that the mates underlying part motions are defined in a consistent way. Specifically, our goal is to ensure maximal connectivity between the parts in an assembly, which is a unique description of its motion. Returning to the telescope example in Figure \ref{fig:dataambiguity}, this means adding an additional mate on the other side of the telescope, since the hinge makes contact on both sides. We can create such maximal connectivity if we ensure that all pairs of parts that can be mated are mated. If we have consistent criteria for which parts can be mated, we can create the additional mates by using the degrees of freedom inferred from the existing mates. Specifically, we examine the chain of existing mates connecting the two parts (which must exist after the previous filter), and define a new mate using the degrees of freedom allowed by the existing mates' combined constraints.  

Our key insight is that we can use geometric cues to identify which pairs of parts can be mated. Namely, we note that 1) in physically realizable assembles, mated parts are in contact and 2) mates are predominantly created from a discrete set of axes derived from the geometry of each part (cylinder axes, face center normals, etc.); we use the set of unique mate coordinate frame z-axes from \citet{jones2021sb}, keeping only the ray direction and offset rather than the full relative coordinate transform, since this is only useful if one needs to derive the part transforms, which we already have. 

We use these two geometric assumptions that determine which pairs of parts can be mated to filter out assemblies in our dataset. First, we discard assemblies with parts that are mated together but do not satisfy the geometric assumption. This discards 36\% of the cleaned assemblies. By visual inspection, we noticed that these models tend to be physically implausible, i.e. they have floating parts. We further discard assemblies for which we could not add all the missing mates---pairs of parts that should be mated under our maximal mate assumption, but whose derived relative motion is not a simple motion type. This discards 14\% of the remaining assemblies. At the end of this process, we are left with a total of 7,328 assemblies in our final set.

\paragraph{Validation Set} Finally, we recruited 4 CAD experts to construct a consistent set of type labels for a sub-collection of assemblies, hand selected by us for visual clarity. We presented the participants with mate-less CAD assemblies with pre-positioned parts, and asked them to add mates using a commercial CAD software. A total of 100 assemblies were annotated 3 times each to generate validated labels by majority consensus.

\section{Motion Prediction}
Our system takes as input an assembly represented as a B-rep describing the various parts. We infer the connectivity structure using the criteria discussed in the previous section, so the remaining task is to find the correct motion type and the correct motion axis, when applicable (see below). 

\paragraph{Motion Type} We used the SBGCN architecture \cite{jones2021sb} to encode and pool the topological features from each part to form axis and part-level features. 
For each pair of mated parts, we combine the resulting features as input to an MLP, which outputs four class probabilities, corresponding to mate types. We experimented with many variations and additions, but found that none made a noticeable improvement. We discuss these further in Section \ref{sec:experiments}. 

\paragraph{Motion Axis} For predicting the correct axis of motion, we select among the possible shared axes between all (touching) pairs of parts in the assembly (see Section \ref{sec:data}). For certain mate types, multiple axes may be equally valid: For sliders, any axis with the same direction is equally valid; for fastens, any axis is valid as there is no motion. Taking this into account, in 88.2\% of mates, there are no incorrect choices for the axis location. For the remaining cases, we train a predictor to infer a probability score for each axis to be used in a mate, and then take the maximum probability axis among each group of axes belonging to the same pair of parts as the location of the mate axis. Similar to the type predictor, we use SBGCN to obtain pooled features for each part, to which we concatenate the features of topological entities used in each axis, which are input to the final MLP layer.

\section{Experimental Results}
\label{sec:experiments}
We split the assemblies into train, validation, and test sets with a split of 80\% - 10\% - 10\%. We created two additional test sets based on the original test set:
\begin{itemize}[noitemsep]
    \item A hand-selected subset of the test set consisting of assemblies that look like a human could infer what they are
    \item A subset of the above assemblies, with the mate labels recreated by consensus of human experts (see the description of the user study in Section \ref{sec:data}.
\end{itemize}
\paragraph{Type Prediction} The accuracy of our mate type predictor is 65\% on the full dataset, 62.8\% on the handpicked data, and 49.4\% on a manually created test set (see below). 
We tried various modifications to our network architecture and features used during learning. As a baseline, we attempted to train a predictor using PointNet \cite{qi2016pointnet} rather than SBGCN, and found that the accuracy on the full dataset dropped from 65\% to 58.5\%. 
Sampling surface points and incorporating the UV-Net encoder of \citet{jayaraman2021uvnet}, then concatenating the resulting features to those of SBGCN makes no difference to the accuracy. We also attempted to incorporate assembly-wide context in various ways, such as adding graph message passing layers between mated parts, and passing a surface point cloud of the entire assembly. These methods fail to make a difference. We also attempted to incorporate axis information in the form of per-topology SBGCN features of the mate connectors along which parts are mated, or point clouds depicting snapshots of rotating and sliding motions between the parts, but it did not help. Finally, we created a heuristic set of labels for each mate, indicating whether it should be able to rotate or slide based on geometric analysis of the part overlaps in motion, and used these to filter our dataset. The accuracy was unaffected by training on this subset, but we note that the test accuracy when restricted to this subset was 72\%, up from 65\%. Using these heuristic labels as additional input did not help, however.

\paragraph{Location Prediction} For the mate axis location problem, we get 71\% accuracy among the mates where there is more than one choice of axis (only 12.8\% of the data), resulting in 96.3\% accuracy overall.


\paragraph{Comparison to Automate} Direct comparison with Automate~\cite{jones2021sb} is not possible since Automate predicts among 8 classes, and has a far more skewed mate type distribution. Instead we compare the accuracy improvement over a model that predicts the most common mate type in each dataset: fastened for Automate, and revolute for ours. Automate achieves roughly a 10\% lift by this metric, whereas our work produces a 25\% improvement.


\subsection{User Study}
Not all mates in the dataset had a corresponding mate chosen by our experts (even after densifying the mates as discussed in Section \ref{sec:data}). Ultimately, we obtained a set of 341 mates for which at least two expert-defined mates could be compared, out of the full 349. Out of these 341 mates, 301 had an agreed upon type according to those two or more experts, so 88\% of mates which could be compared between multiple experts were agreed to be of one particular type. The type chosen by consensus among these mates agreed with the original mate type 66.8\% of the time. 
On average, in 68.8\% of the mates, the original type in the dataset agreed with the mate types the experts chose chose. 

\section{Conclusions}
Our performance on the hand-picked data is slightly worse than on the full data, indicating that the subjective criteria by which we deemed those assemblies reasonable do not make them easier to predict for our machine learning model. Furthermore, the performance is much worse on the user study-based test set, which might suggest that the task of inserting missing mates into an assembly results in different biases in assigning mate types than assembling from scratch, or that some CAD assemblies are not annotated with functional motion in mind. The agreement rate of the expert labels, both with each other and with the original mates, suggests that human performance is bounded at about 70\%. We believe that the performance achieved in this work can be improved upon, and that our expert-generated test dataset will assist with evaluating future works on motion prediction.

\section{Acknowledgements}
This work was funded by the UW Reality Lab, Meta, Google, OPPO, and Amazon.

\printbibliography

\end{document}